%% file: ijcai22.tex
\documentclass{article}
\usepackage{ijcai22}

\usepackage{times}
\usepackage{soul}
\usepackage{url}
\usepackage[hidelinks]{hyperref}
\usepackage[utf8]{inputenc}
\usepackage[small]{caption}
\usepackage{graphicx}
\usepackage{amsmath}
\usepackage{amsthm}
\usepackage{booktabs}
\usepackage{algorithm}
\usepackage{algorithmic}
\usepackage{dsfont}
\usepackage{multirow}
\usepackage{bbding}
\usepackage{pifont}
\newcommand{\cmark}{\ding{51}}%
\newcommand{\xmark}{\ding{55}}%
\usepackage{color}
\usepackage{colortbl}
\definecolor{c_gray}{gray}{.9}
\usepackage[ruled,noend,algo2e]{algorithm2e}

\title{FQ-ViT: Post-Training Quantization for Fully Quantized Vision Transformer}

\author{
Yang Lin\footnotemark[1]\footnotemark[2]\and
Tianyu Zhang\footnotemark[1] \and
Peiqin Sun\footnotemark[3]\and
Zheng Li \And
Shuchang Zhou\\
\affiliations
MEGVII Technology\\
\emails
linyang.zhh@gmail.com, \{zhangtianyu, sunpeiqin, lizheng02, zsc\}@megvii.com
}

\begin{document}

\maketitle

\renewcommand{\thefootnote}{\fnsymbol{footnote}}
\footnotetext[1]{Equal contribution.}
\footnotetext[2]{Work done while interning at MEGVII Technology.}
\footnotetext[3]{Corresponding author.}

\begin{abstract}
Network quantization significantly reduces model inference complexity and has been widely used in real-world deployments. However, most existing quantization methods have been developed mainly on Convolutional Neural Networks~(CNNs), and suffer severe degradation when applied to fully quantized vision transformers. In this work, we demonstrate that many of these difficulties arise because of serious inter-channel variation in LayerNorm inputs, and present, Power-of-Two Factor~(PTF), a systematic method to reduce the performance degradation and inference complexity of fully quantized vision transformers. In addition, observing an extreme non-uniform distribution in attention maps, we propose Log-Int-Softmax~(LIS) to sustain that and simplify inference by using 4-bit quantization and the BitShift operator. Comprehensive experiments on various transformer-based architectures and benchmarks show that our Fully Quantized Vision Transformer~(FQ-ViT) outperforms previous works while even using lower bit-width on attention maps. For instance, we reach 84.89\% top-1 accuracy with ViT-L on ImageNet and 50.8 mAP with Cascade Mask R-CNN~(Swin-S) on COCO. To our knowledge, we are the first to achieve lossless accuracy degradation~($\sim$1\%) on fully quantized vision transformers. The code is available at \url{https://github.com/megvii-research/FQ-ViT}.
\end{abstract}

\section{Introduction}

Transformer-based architectures have achieved competitive performance in various computer vision~(CV) tasks, including image classification~\cite{dosovitskiy2021an,touvron2021training}, object detection~\cite{carion2020end,liu2021swin}, semantic segmentation~\cite{zheng2021rethinking} and so on. Compared to the CNN counterparts, transformers usually have more parameters and higher computational costs. For example, ViT-L has 307M parameters and 190.7G FLOPs, reaching the accuracy of 87.76\% in ImageNet with large-scale pre-training. However, the large number of parameters and computational overhead of transformer-based architectures present a challenge when deployed to resource-constrained hardware devices.

To facilitate deployment, several techniques have been proposed, including quantization~\cite{zhou2016dorefa,nagel2020up,shen2020q,liu2021post}, pruning~\cite{han2015deep}, distillation~\cite{jiao2019tinybert} and adaptation of architecture design~\cite{graham2021levit}. We focus on the quantization technique in this paper and note that pruning, distillation, and architecture adaptation are orthogonal to our work and can be combined.

\begin{figure}[t]

\centering
\includegraphics[width=8.1cm]{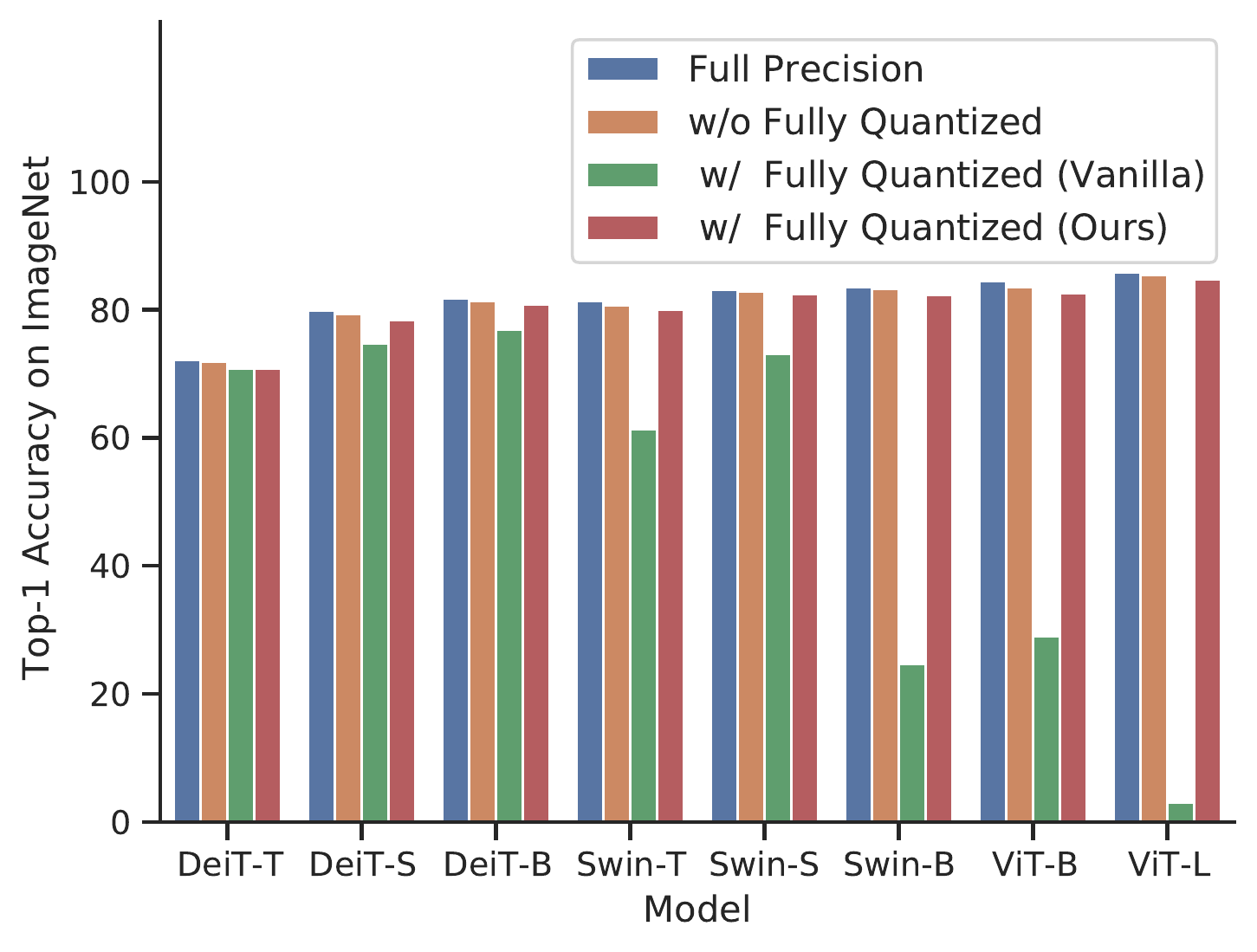}
\vskip -0.1in
\caption{Top-1 accuracy on ImageNet for full precision and quantized vision transformers.
w/o Fully Quantized: LayerNorm and Softmax remain floating-point, while other modules are quantized to 8-bit by MinMax.
w/ Fully Quantized~(Vanilla): all modules are quantized to 8-bit by MinMax.
w/ Fully Quantized~(Ours): all modules are quantized to 8-bit by our method. 
}
\label{fig:intro}
\vskip -0.1in
\end{figure}

Most existing quantization approaches have been designed and tested on CNNs and lack proper handling of transformer-specific constructs. Previous work~\cite{liu2021post} finds there have been a significant accuracy degradation when quantizing LayerNorm and Softmax of vision transformers. In this case, the models are not fully quantized, resulting in the need to retain floating-point units in the hardware, which will bring large consumption and significantly reduce the inference speed~\cite{8716697}.
So we revisit these two exclusive modules of the vision transformers and discover the reasons of degradation. Firstly, we find a serious inter-channel variation of LayerNorm inputs, which some channel ranges even exceed 40$\times$ of the median. Traditional methods cannot handle such large fluctuations of activations, which will lead to large quantization error.
Secondly, we find that the values of the attention map have an extreme non-uniform distribution, with most values clustered in 0 $\sim$ 0.01, and a few high attention values close to 1.

Based on the analysis above, we propose Power-of-Two Factor~(PTF) to quantize the inputs of the LayerNorm. In this way, the quantization error is greatly reduced, and the overall computational efficiency is the same as that of layer-wise quantization thanks to the BitShift operator.
In addition, we propose Log-Int-Softmax~(LIS), which provides higher quantization resolution for small values and presents a more efficient integer inference for Softmax.
Combining these methods, we are the first work to achieve post-training quantization for fully quantized vision transformers. As shown in Figure~\ref{fig:intro}, our method significantly improves the performance of fully quantized vision transformers and obtains comparable accuracy with full precision counterparts.

Our contributions are four-fold:
\begin{itemize}
    \item We revisit the fully quantized vision transformers and attribute the accuracy degradation to the serious inter-channel variation in LayerNorm inputs. Meanwhile, we observe an extreme non-uniform distribution of attention maps, resulting in another part of the quantization error.
    \item We propose Power-of-Two Factor~(PTF), a simple yet efficient post-training method that can achieve accurate quantization on LayerNorm inputs with only one layer-wise quantization scale.
    \item We propose Log-Int-Softmax~(LIS), a novel method that can perform 4-bit quantization on attention maps. With LIS, we can store attention maps on an aggressively low-bit and replace multiplication with BitShift operator. We achieve integer-only inference on Softmax modules, significantly reducing the inference consumption.
    \item We conduct extensive experiments on image classification and object detection with various transformer-based architectures. The results show that our fully quantized vision transformers with 8-bit weights/activations and 4-bit attention maps, can achieve comparable performance to floating-point versions.
\end{itemize}

\section{Related Work}
\subsection{Vision Transformer}
Recently, transformer-based architecture shows great power in CV tasks. Emerging works based on ViT~\cite{dosovitskiy2021an} demonstrated the effectiveness across all vision tasks such as classification~\cite{touvron2021training}, detection~\cite{carion2020end} and segmentation~\cite{zheng2021rethinking}.
The newly proposed Swin Transformer~\cite{liu2021swin} even surpasses the state-of-the-art CNNs on almost tranditional CV tasks, presenting strong expressive and generalization capability of transformer.
However, these high-performing vision transformers are attributed to the large number of parameters and high computational overhead, limiting their adoption.
Therefore, innovating a smaller and faster vision transformer becomes a new trend.
Le-ViT~\cite{graham2021levit} makes progress in faster inference with down-sampling, patch descriptors, and a redesign of Attention-MLP block. DynamicViT~\cite{rao2021dynamicvit} presents a dynamic token sparsification framework to prune redundant tokens progressively and dynamically, achieving competitive complexity and accuracy trade-off. 
Evo-ViT~\cite{xu2021evo} proposes a slow-fast updating mechanism that guarantees information flow and spatial structure, trimming down both the training and inference complexity.
While the above works focus on efficient model designing, this paper boosts the compression and acceleration in the track of quantization.

\begin{figure*}[t]
\centering
\includegraphics[width=17.0cm]{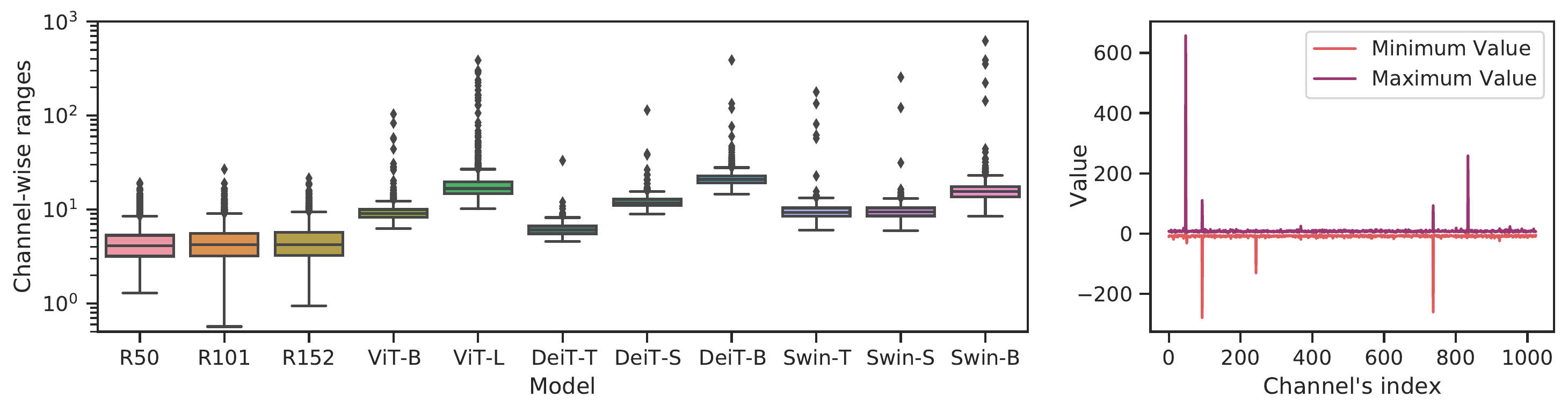}
\caption{\textbf{Left}: Boxplot of the last LayerNorm inputs' channel-wise ranges in each model. \textbf{Right}: Channel-wise minimum and maximum values of the last LayerNorm inputs in full precision Swin-B. The above two figures show that there exists more serious inter-channel variation in vision transformers than CNNs, which leads to unacceptable quantization errors with layer-wise quantization.}
\vskip -0.07in
\label{fig:range}
\end{figure*}

\subsection{Network Quantization}
Current quantization methods can be divided into two categories: Quantization-Aware Training~(QAT) and Post-Training Quantization~(PTQ). QAT~\cite{zhou2016dorefa,jacob2018quantization} depends on training to achieve aggressively low-bit~(e.g. 2-bit) quantization and promising performance, 
while it often requires a high-level expert knowledge and huge GPU resources for training or fine-tuning.
To reduce above costs of quantization, PTQ, which is training-free, has received more widespread attention and lots of excellent works arise. OMSE~\cite{choukroun2019low} proposes to determine the value range of activation by minimizing the quantization error. AdaRound~\cite{nagel2020up} presents a novel rounding mechanism to adapt the data and the task loss. 
Besides works above specific to CNNs, \citeauthor{liu2021post} proposes a post-training quantization method for vision transformers with similarity-aware and rank-aware strategies. However, this work does not quantize Softmax and LayerNorm modules, resulting in an incomplete quantization. In our FQ-ViT, we aim to implement an accurate, fully quantized vision transformer under the PTQ paradigm.

\section{Proposed Method}
In this section, we will introduce our proposed approach in detail. First in Section~\ref{sec:method_pre}, we present the preliminary of network quantization. Then in Section~\ref{sec:ptf} and~\ref{sec:lis}, we analyze the reasons of degradation in fully quantized vision transformers and propose two novel quantization methods, Power-of-Two Factor~(PTF) and Log-Int-Softmax~(LIS), for LayerNorm and Softmax.

\subsection{Preliminary}
\label{sec:method_pre}

In this section, we explain the notations of network quantization.
Assuming the quantization bit-width is $b$, the quantizer $\textrm{Q}(\textrm{X}|b)$ can be formulated as a function that maps a floating-point number $\textrm{X}\in \mathds{R}$ to the nearest quantization bin:

\begin{equation}
    \textrm{Q}(\textrm{X}|b): \mathds{R} \rightarrow \textrm{q},
\end{equation}
\begin{equation}
\textrm{q}=
\left\{\begin{aligned}
&\{-\textrm{2}^{b-1},\cdots ,\textrm{2}^{b-1}-\textrm{1}\} &Signed,\\
&\{\textrm{0}, \textrm{1} \cdots ,\textrm{2}^{b}-\textrm{1}\}  &Unsigned.
\end{aligned}\right.
\end{equation}

There are various quantizer $\textrm{Q}(\textrm{X}|b)$, where uniform~\cite{jacob2018quantization} and log2~\cite{cai2018deep} are typically used.

\textbf{Uniform Quantization} is well supported on most hardware platforms. Its quantizer $\textrm{Q}(\textrm{X}|b)$ can be defined as:

\begin{equation}
    \textrm{Q}(\textrm{X}|b)=\operatorname{clip}(\lfloor\frac{\textrm{X}}{s}\rceil+zp, \textrm{0}, \textrm{2}^{b}-\textrm{1}),
\end{equation}
where $s$~(scale) and $zp$~(zero-point) are quantization parameters determined by the lower bound $l$ and the upper bound $u$ of $\textrm{X}$, which are usually minimum and maximum values:
\begin{align}
l=&\min(\textrm{X}), u=\max(\textrm{X}),\\
    s=&\frac{u-l}{\textrm{2}^b-\textrm{1}},zp=\operatorname{clip}(\lfloor-\frac{l}{s}\rceil, \textrm{0}, \textrm{2}^{b}-\textrm{1}).
\end{align}

\textbf{Log2 Quantization} converts the quantization process from linear to exponential variation. Its quantizer $\textrm{Q}(\textrm{X}|b)$ can be defined as: 
\begin{equation}
    \textrm{Q}(\textrm{X}|b)=\operatorname{sign}(\textrm{X})\cdot \operatorname{clip}(\lfloor -\log_\textrm{2}\frac{|\textrm{X}|}{\max(|\textrm{X}|)}\rceil, \textrm{0},\textrm{2}^{b-1}-\textrm{1}).
    \label{equ:log2}
\end{equation}

In this paper, to achieve a fully quantized vision transformer, we quantize all modules, including Conv, Linear, MatMul, LayerNorm, Softmax, etc.
Especially, uniform MinMax quantization is used for Conv, Linear and MatMul modules and our following methods are used for LayerNorm and Softmax.

\subsection{Power-of-Two Factor for LayerNorm Quantization}
\label{sec:ptf}

During inference, LayerNorm~\cite{ba2016layer} computes the statistics $\mu_\textrm{X}$,$\sigma_\textrm{X}$ in each forward step and normalizes input $\textrm{X}$. Then, affine parameters $\gamma$, $\beta$ rescale the normalized input to another learned distribution. The above process can be written as:
\begin{equation}
    \operatorname{LayerNorm}(\textrm{X}) = \frac{\textrm{X}-\mu_\textrm{X}}{\sqrt{\sigma_\textrm{X}^2+\epsilon}}\cdot\gamma + \beta.
\end{equation}

Unlike BatchNorm~\cite{ioffe2015batch}, commonly used in CNNs, LayerNorm cannot be folded into the previous layer due to its dynamic computational property, so we have to quantize it separately. However, we observe a significant performance degradation while applying post-training quantization on it.
Looking into the inputs of LayerNorm layers, we find there is a serious inter-channel variation.
Figure~\ref{fig:range} presents the channel-wise ranges of activation in the last LayerNorm layer. In addition, we also display the cases of ResNets~\cite{he2016deep} for comparison. Considering that there is no LayerNorm in ResNets, we choose the activations at the same position~(fourth stage's outputs) to exhibit.

It is observed that the channel-wise ranges fluctuate more wildly in vision transformers than those in ResNets. For instance, the maximum range/median range of ResNet152 is only 21.6/4.2, while it goes up to 622.5/15.5 in Swin-B.
Based on such extreme inter-channel variation, layer-wise quantization, which applies the same quantization parameters to all channels, will lead to an intolerable quantization error. A possible solution is using group-wise quantization~\cite{shen2020q} or channel-wise quantization~\cite{li2019fully}, which assign different quantization parameter to different group or channel. However, these will still induce the calculation of mean and variance in the floating-point domain, resulting in a high hardware overhead.

In this paper, we propose a simple yet efficient method, Power-of-Two Factor~(PTF), for LayerNorm quantization. The core idea of PTF is to equip different channels with different factors, rather than different quantization parameters. Given the quantization bit-width $b$, the input activation $\textrm{X}\in \mathds{R}^{\textrm{B}\times \textrm{L}\times \textrm{C}}$, the layer-wise quantization parameters $s,zp\in \mathds{R}^\textrm{1}$, and the PTF $\alpha \in \mathds{N}^\textrm{C}$, then the quantized activation $\textrm{X}_\textrm{Q}$ can be formulated as:
\begin{equation}
    \textrm{X}_\textrm{Q}=\textrm{Q}(\textrm{X}|b)=\operatorname{clip}(\lfloor\frac{\textrm{X}}{\textrm{2}^{\alpha}s}\rceil+zp, \textrm{0},\textrm{2}^{b}-1),
\end{equation}
with
\begin{equation}
    s=\frac{\max(\textrm{X})-\min(\textrm{X})}{\textrm{2}^b-\textrm{1}}~/~\textrm{2}^\textrm{K},
\end{equation}
\begin{equation}
    zp=\operatorname{clip}(\lfloor-\frac{\min(\textrm{X})}{\textrm{2}^\textrm{K}s}\rceil, \textrm{0}, \textrm{2}^{b}-\textrm{1}),
\end{equation}
\begin{equation}
    \alpha_c=\underset{{\alpha_c\in\{\textrm{0},\textrm{1},\cdots,\textrm{K}\}}}{\operatorname{\arg \min}}\left\|\textrm{X}_c-\lfloor\frac{\textrm{X}_c}{\textrm{2}^{\alpha_c}s}\rceil\cdot \textrm{2}^{\alpha_c}s\right\|_\textrm{2}.
\end{equation}
Noticing $c$ represents the channel index for $\textrm{X}$ and $\alpha$. The hyperparameter $\textrm{K}$ could meet different scaling requirements. In order to cover the different inter-channel variation across all models, we set $\textrm{K}=\textrm{3}$ as default. Detailed experiments can be found in supplementary materials.

At this point, each channel has its own Power-of-Two Factor $\alpha$ and layer-wise parameters $s$,$zp$. 
During inference, layer-wise parameters $s$ and $zp$ can be extracted, so the computation of $\mu$,$\sigma$ could be done in the integer domain rather than floating-point, which reduces the energy and area costs~\cite{8716697}. Meanwhile, thanks to the nature of powers of two, PTF $\alpha$ can be efficiently combined with layer-wise quantization by BitShift operator, avoiding floating-point calculations of 
group-wise or channel-wise quantization. The whole process can be processed with two phases:

Phase 1: Shift the quantized activation with Power-of-Two Factor $\alpha$:
\begin{equation}
    \widehat{\textrm{X}}_\textrm{Q}=(\textrm{X}_\textrm{Q}-zp)<<\alpha.
\end{equation}

Phase 2: Calculate the mean and variance based on the shifted activation $\widehat{\textrm{X}}_\textrm{Q}$:
\begin{align}
    \mu(\textrm{X})&\approx\mu(\textrm{2}^\alpha s\cdot (\textrm{X}_\textrm{Q}-zp))=s\cdot \mu(\widehat{\textrm{X}}_\textrm{Q}),\\
    \sigma(\textrm{X})&\approx\sigma(\textrm{2}^\alpha s\cdot (\textrm{X}_\textrm{Q}-zp))=s\cdot\sigma(\widehat{\textrm{X}}_\textrm{Q}).
\end{align}

\begin{figure}[t]
\centering
\includegraphics[width=8.3cm]{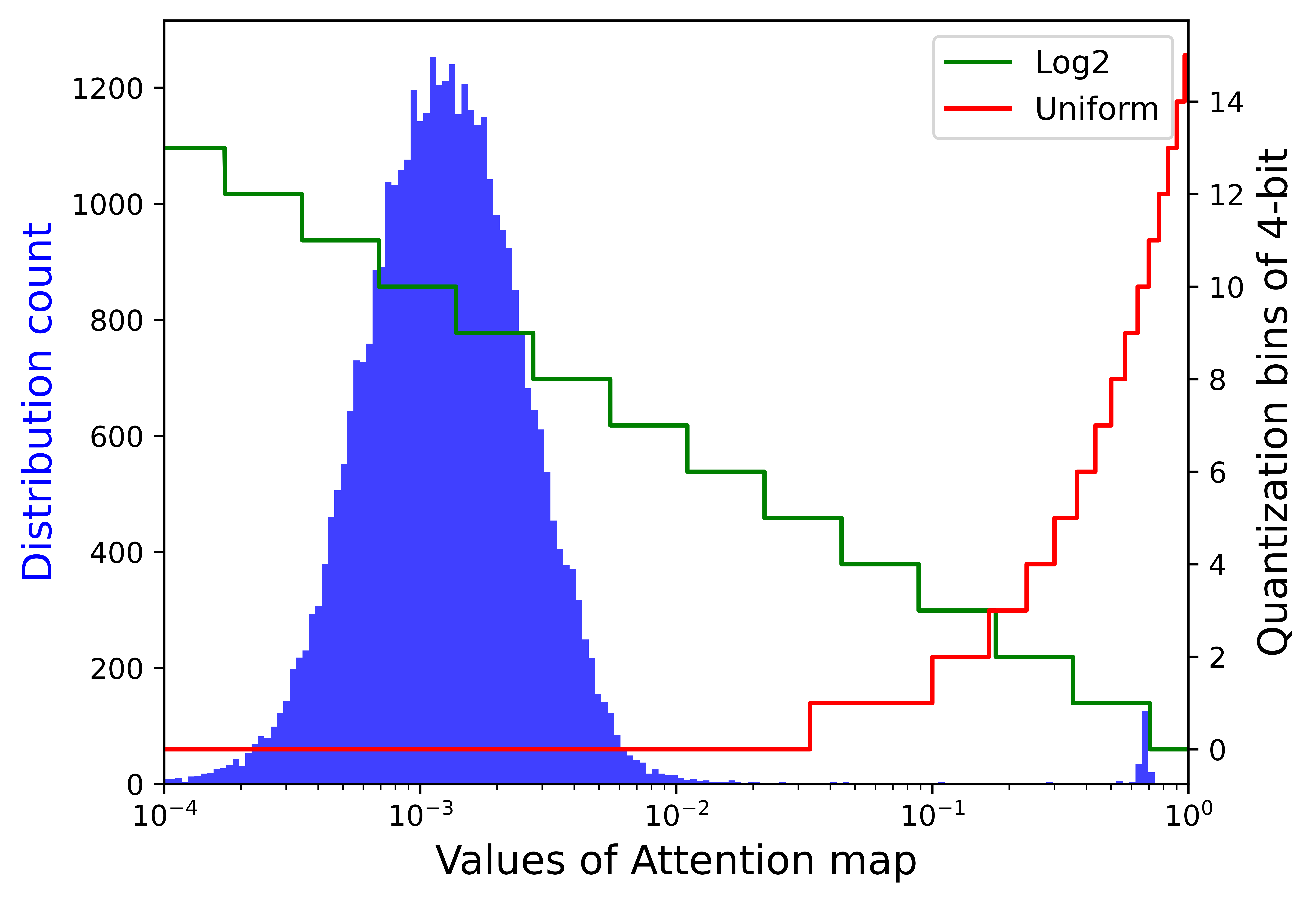}
\caption{Distribution of attention maps in ViT-L with visualizing the 4-bit quantized bins of uniform and log2 quantization. X-axis is in log-scale and we can observe that log2 quantization preserves more bins than uniform for small values.}
\label{fig:sm_dis} 
\end{figure}

\subsection{Log-Int-Softmax for Softmax Quantization}
\label{sec:lis}

Multi-head Self-Attention~(MSA) is one of the most important components in transformer-based architectures, while it is considered the most resource-intensive due to the quadratic complexity to the number of token, the division of the image resolution by the patch size. As model performance proved to benefit from higher resolution and smaller patch size~\cite{dosovitskiy2021an}, when the increasing resolution and reducing patch size, the storage and computation of attention maps become the bottleneck which directly affect the throughput and latency of inference. Therefore, smaller attention maps and more efficient inference become an urgent need.

\subsubsection{Log2 Quantization for Attention Map}

In order to compress the attention maps to smaller size and speed up the inference, we quantize attention maps to lower bit-width. As experimenting on quantization of attention maps from 8-bit to 4-bit with uniform quantization, all vision transformers show severe performance drop. For example, DeiT-T only results in 8.69\% top-1 accuracy on ImageNet with 4-bit uniform quantized attention maps, decreasing 63.05\% from 8-bit case.

\begin{figure}[t]
\centering
\includegraphics[width=8.55cm]{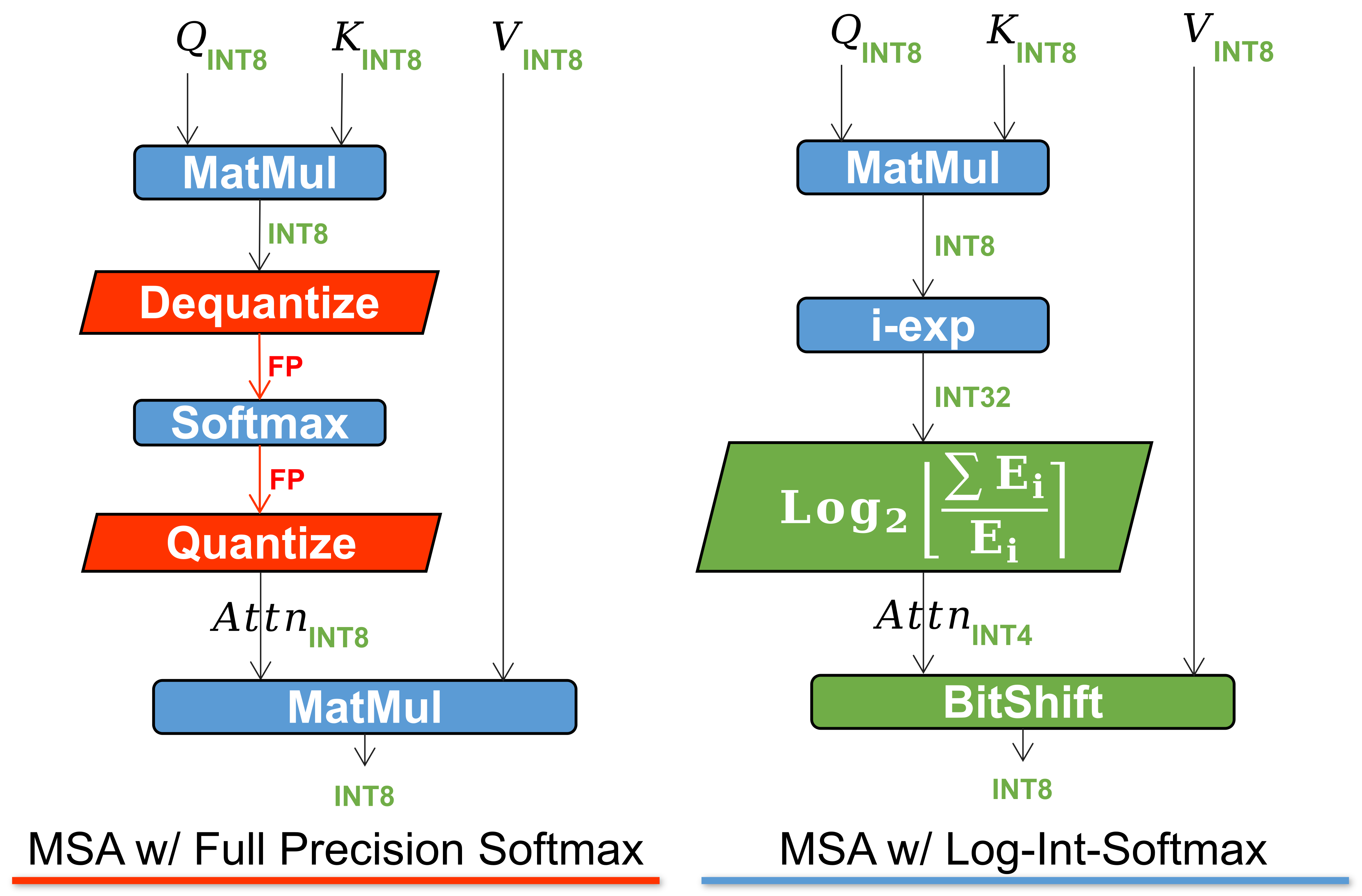}
\vskip 0.03in
\caption{Comparison of using full precision Softmax and Log-Int-Softmax in quantized multi-head self-attention inference. Full precision Softmax needs to dequantize and requantize around Softmax, while LIS keeps an integer-only data type in the whole MSA inference.}
\label{fig:com_sm}
\end{figure}

Inspired by the idea of sparse attention in DynamicViT~\cite{rao2021dynamicvit}, we probe into the distribution of attention maps, as Figure~\ref{fig:sm_dis} shows. We observe a distribution centering at a fairly small value, while only a few outliers have larger values close to 1. Averaging on all attention maps in ViT-L, about 98.8\% of values are smaller than $1/16$. Compared with 4-bit uniform quantization which only assigns 1 bin for such many values, log2 method can allocate 12 bins for them. Moreover, following the purpose of ranking-aware loss~\cite{liu2021post}, log2 quantization can retain much order consistency between full precision and quantized attention maps. Consequently, we save the extreme degradation in 4-bit quantization of attention maps and achieve equal performance as 8-bit uniform quantization with 50\% less memory footprint.

\begin{table*}[t]
\centering
\small
\begin{tabular}{lrrrrrrrrr}
\toprule
Method            & W/A/Attn & DeiT-T         & DeiT-S         & DeiT-B         & ViT-B          & ViT-L          & Swin-T         & Swin-S         & Swin-B         \\
\midrule
\rowcolor{c_gray} Full Precision    & 32/32/32      & 72.21          & 79.85          & 81.85          & 84.53          & 85.81          & 81.35          & 83.20          & 83.60          \\
\midrule
MinMax            & 8/8/8         & 70.94          & 75.05          & 78.02          & 23.64          & 3.37           & 64.38          & 74.37          & 25.58          \\
EMA~\cite{jacob2018quantization}               & 8/8/8         & 71.17          & 75.71          & 78.82          & 30.30          & 3.53           & 70.81          & 75.05          & 28.00          \\
Percentile~\cite{li2019fully}        & 8/8/8         & 71.47          & 76.57          & 78.37          & 46.69          & 5.85           & 78.78          & 78.12          & 40.93          \\
OMSE~\cite{choukroun2019low}              & 8/8/8         & 71.30          & 75.03          & 79.57          & 73.39          & 11.32          & 79.30          & 78.96          & 48.55          \\
Bit-Split$^*$~\cite{wang2020towards}         & 8/8/8         & -              & 77.06          & 79.42          & -              & -              & -              & -              & -              \\
PTQ for ViT$^*$~\cite{liu2021post}       & 8/8/8         & -              & 77.47          & 80.48          & -              & -              & -              & -              & -              \\
\midrule
\multirow{2}{*}{FQ-ViT}      & 8/8/8         & \textbf{71.61} & \textbf{79.17} & \textbf{81.20} & \textbf{83.31} & \textbf{85.03} & \textbf{80.51} & \textbf{82.71} & \textbf{82.97} \\
 & 8/8/4         & \textbf{71.07} & \textbf{78.40} & \textbf{80.85} & \textbf{82.68} & \textbf{84.89} & \textbf{80.04} & \textbf{82.47} & \textbf{82.38} \\
\bottomrule
\end{tabular}
\caption{Comparison of the top-1 accuracy with state-of-the-art methods on ImageNet dataset. $^*$ indicates that all LayerNorm and Softmax modules are not quantized.}
\label{tab:img}
\end{table*}

Log2 quantization proved to be suitable combining with MSA from two aspects. Firstly, comparing to Equation~(\ref{equ:log2}), the fixed output range $(\textrm{0},\textrm{1})$ of Softmax makes the log2 function calibration-free:
\begin{equation}
     \textrm{Attn}_\textrm{Q} = \textrm{Q}(\textrm{Attn}|b) = \operatorname{clip}(\lfloor-\log_\textrm{2}(\textrm{Attn})\rceil, \textrm{0}, \textrm{2}^b-\textrm{1}).
\end{equation}
This ensures that the quantization of attention maps will not be affected by the fluctuation of calibration data.

Secondly, it also introduces the merits of converting the MatMul to BitShift between the quantized attention map~$(\textrm{Attn}_\textrm{Q})$ and values~$(\textrm{V}_\textrm{Q})$ as:
\begin{align}
    \textrm{Attn}\cdot \textrm{V}_\textrm{Q} &= \textrm{2}^{-\textrm{Attn}_\textrm{Q}} \cdot \textrm{V}_\textrm{Q}
    = \textrm{V}_\textrm{Q}>>\textrm{Attn}_\textrm{Q} \\ &= \frac{\textrm{1}}{\textrm{2}^N} \cdot (\textrm{V}_\textrm{Q}<<(\textrm{N}-\textrm{Attn}_\textrm{Q})),
    \label{equ:bitshift_attn}
\end{align}
with $\textrm{N} = \textrm{2}^b-\textrm{1}$.
Noticing that directly right shift $\textrm{V}_\textrm{Q}$ with the results of $\textrm{Attn}_\textrm{Q}$ may lead to severe truncation error. We use $(\textrm{N}-\textrm{Attn}_\textrm{Q})$ as the quantized output with a scale equaling $\textrm{1}/{\textrm{2}^N}$, which ensures a left-shift operation to prevent from truncation error.

\subsubsection{Integer-only Inference}

Previous works~\cite{liu2021post} chose to not quantize Softmax because the negligibility of calculation amount in Softmax, and quantization may lead to significant accuracy degradation. However, data moving between CPU and GPU/NPU, doing dequantization and requantization, will induce great difficulties in hardware design, which is not a negligible consumption.

Combining log2 quantization with $\operatorname{i-exp}$~\cite{kim2021bert}, which is a polynomial approximation of exponential function, we propose Log-Int-Softmax, an integer-only, faster, low consuming Softmax:
\begin{equation}
    \exp(s\cdot \textrm{X}_\textrm{Q}) \approx s'\cdot\operatorname{i-exp}(\textrm{X}_\textrm{Q}),
\end{equation}
\begin{equation}
    \operatorname{LIS}(s\cdot \textrm{X}_\textrm{Q}) = \textrm{N} -\log_\textrm{2}\lfloor\frac{\sum \operatorname{i-exp}(\textrm{X}_\textrm{Q})}{\operatorname{i-exp}(\textrm{X}_\textrm{Q})}\rceil,
\end{equation}
with $\textrm{N} = \textrm{2}^b-\textrm{1}$. An integer log2 function can be easily implemented by using BitShift to find the first bit index where the value is 1~(we call it $\operatorname{Find\_First\_One}$ function), and adding the value of bit right behind that. Detailed derivation can be found in the supplementary materials.

The difference between normal MSA and our method are shown in Figure~\ref{fig:com_sm}, with the data type of every stage labeled. In the multi-head self-attention with unquantized Softmax shown on the left, the matrix multiplication of queries~$(\textrm{Q})$ and keys~$(\textrm{K})$ needs to be dequantized to full precision before Softmax, and requantized after it. When our Log-Int-Softmax adopted, shown on the right, the entire data type can be in pure integer, with quantization scale individually and paralleling calculated. 
It is worth noting that LIS uses an aggressively 4-bit representation on attention maps, which significantly reduces memory footprint.

\section{Experiments}

\begin{table*}[t]
\centering
\small
\begin{tabular}{lrrr}
\toprule
\multirow{2}{*}{Method}  & \multirow{2}{*}{W/A/Attn}    & Mask R-CNN & Cascade Mask R-CNN \\
& & w/ Swin-S& w/ Swin-S\\
\midrule
\rowcolor{c_gray} Full Precision    & 32/32/32      & 48.5                & 52.0                   \\
\midrule
MinMax            & 8/8/8         & 32.8                & 35.2                   \\
EMA~\cite{jacob2018quantization}               & 8/8/8         & 37.9                & 40.4                   \\
Percentile~\cite{li2019fully}        & 8/8/8         & 41.6                & 44.7                   \\
OMSE~\cite{choukroun2019low}              & 8/8/8         & 42.6                & 44.9                   \\
\midrule
\multirow{2}{*}{FQ-ViT}      & 8/8/8         & \textbf{47.8}       & \textbf{51.4}          \\
 & 8/8/4         & \textbf{47.2}       & \textbf{50.8}  \\
\bottomrule
\end{tabular}
\caption{Comparison of the bbox mAP with state-of-the-art methods on COCO dataset.}
\label{tab:coco}
\end{table*}

In this section, we present experimental results on vision transformers for image classification and object detection. We state the detailed experimental configuration firstly and then exhibit the comparison of our method with existing post-training quantization methods in ImageNet~\cite{krizhevsky2012imagenet} and COCO~\cite{lin2014microsoft} benchmarks. In the end, ablation studies are conducted to evaluate the effectiveness of Power-of-Two Factor~(PTF) and Log-Int-Softmax~(LIS).

\subsection{Implementation Details}

We randomly sample 1000 training images from ImageNet or COCO as the calibration data, and use the validation set to evaluate performance.
Apart from special notes, we perform symmetric channel-wise quantization for weights and asymmetric layer-wise quantization for activations. For a fair comparison, the quantization for weights is fixed as MinMax.
The hyperparameter $\textrm{K}$ in Power-of-Two Factor is set to 3.

\subsection{Comparison with State-of-the-art Methods}
This paper employs several current post-training quantization methods, including MinMax, EMA~\cite{jacob2018quantization}, Percentile~\cite{li2019fully}, OMSE~\cite{choukroun2019low}, Bit-Split~\cite{wang2020towards} and PTQ for ViT~\cite{liu2021post}.
Note that PTQ for ViT~\cite{liu2021post} is closest to our work, however it does not quantize the LayerNorm and Softmax, while we quantize all modules.

\subsubsection{Image Classification on ImageNet}

To demonstrate the effectiveness of proposed methods, we conduct extensive experiments in ImageNet~\cite{krizhevsky2012imagenet} with various vision transformers, i.e., ViT~\cite{dosovitskiy2021an}, DeiT~\cite{touvron2021training}, Swin Transformer~\cite{liu2021swin}.
The overall top-1 accuracy results are reported in Table~\ref{tab:img}.
It is obvious that all current methods can’t capture fully quantized vision transformers, while our FQ-ViT does it and achieves a nearly lossless quantization even with an aggressively low-bit on attention maps. Meanwhile, our FQ-ViT significantly exceeds PTQ for ViT~\cite{liu2021post}, whose LayerNorm and Softmax are not quantized. For instance, our FQ-ViT achieves 81.20\% accuracy on DeiT-B in the case of all modules quantized to 8-bit, and it can still achieve 80.85\% accuracy when the attention maps are compressed to 4-bit.

\subsubsection{Object Detection on COCO}

We also conduct experiments on the object detection benchmark COCO~\cite{lin2014microsoft}. We choose Swin series~\cite{liu2021swin} detectors for experiments and the results can be found in Table~\ref{tab:coco}. It is observed that all current methods have poor performance on fully quantized detectors. 
Our FQ-ViT  significantly improves the quantization accuracy and achieves 47.2 mAP on Mask R-CNN~(Swin-S) and 50.8 mAP on Cascade Mask R-CNN~(Swin-S) with 8-bit on weights/activations and 4-bit on attention maps.

\begin{table}[t]
\centering
\small
\begin{tabular}{lccrr}
\toprule
Method & PTF & LIS & BitOPs~(G) & Acc.~(\%)\\
\midrule
\rowcolor{c_gray} Full Precision   & -                   & -                 & -       &         84.53        \\
\midrule
Baseline \#8                  & \xmark                   & \xmark                 &  1118.51     &          23.64       \\
    & \cmark                   & \xmark                &      1118.52                     &  \textbf{83.31} \\
    \midrule
Baseline \#4     & \cmark                   & \xmark                &       1118.34                     &        7.96         \\
    &\cmark                  & \cmark                 &              1117.76              &          \textbf{82.68}      \\

\bottomrule
\end{tabular}
\caption{Effect of the Power-of-Two Factor~(PTF) and Log-Int-Softmax~(LIS). We evaluate the performance of full precision and quantized ViT-B on ImageNet validation set. We choose MinMax with 8-bit weights and activations as "Baseline" and \# indices the bit-width of attention maps.}
\label{tab:ab}
\end{table}
\subsection{Ablation Studies}

\begin{figure}[t]
\centering
\includegraphics[width=8.3cm]{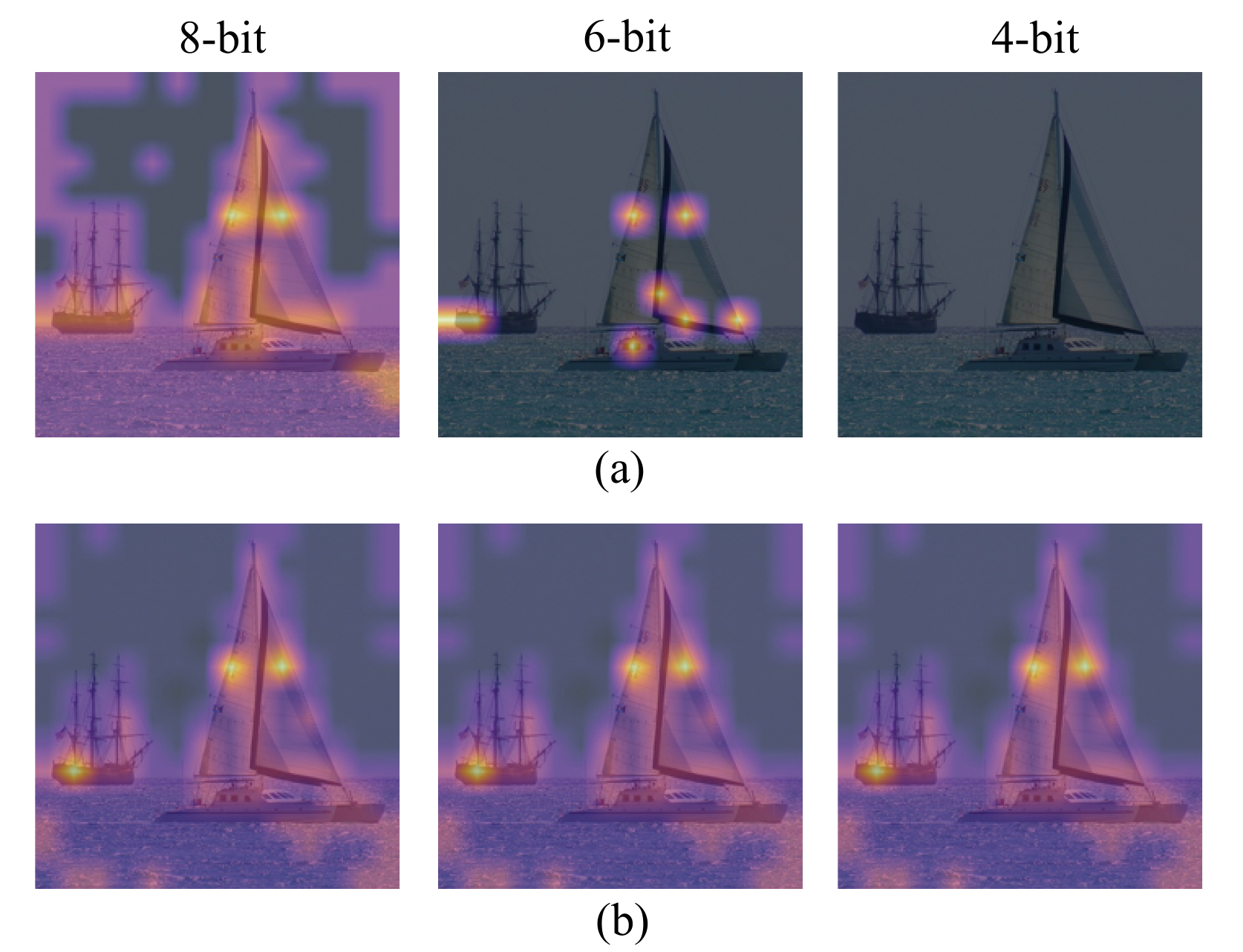}
\caption{Attention map visualization. (a) shows the results of uniform quantization and (b) shows the results of our Log-Int-Softmax.}
\label{fig:attn_vis}
\vskip 0.1in
\end{figure}

To study the effect of our methods, Power-of-Two Factor~(PTF) and Log-Int-Softmax~(LIS), we fully quantize ViT-B under a variety of strategies and report the results in Table~\ref{tab:ab}. We design two baselines. "Baseline \#8" indices the model is fully quantized by MinMax with 8-bit weights, activations and attention maps, while "Baseline \#4" has a lower bit-width~(4-bit) on attention maps. From the results, we have several observations. Firstly, the model with PTF and LIS outperforms the baseline and achieves almost lossless accuracy. Secondly, thanks to the BitShift operator, PTF only introduces little amount of BitOPs, while LIS reduces that.

\subsection{Visualization of Quantized Attention Map}
We visualize the attention maps to see the difference between uniform quantization and our LIS as Figure~\ref{fig:attn_vis} shows. When both using 8-bit, uniform quantization focuses on the high activation area, while LIS keeps more texture in the low activation area, which retains more relative rank of the attention map. This divergence does not make a big difference in the case of 8-bit. However, when quantized to lower bit-width, as the 6-bit and 4-bit cases show, uniform quantization sharply degrades and even deactivates all the attention area. On the contrary, LIS still presents acceptable performance similar to 8-bit.

\section{Conclusions}
In this paper, we propose a method to fully quantize vision transformers. Specifically, we propose Power-of-Two Factor~(PTF) to deal with the serious inter-channel variations in the inputs of LayerNorm. In addition, we propose an integrated quantization solution Log-Int-Softmax~(LIS) to implement 4-bit quantization of attention maps and utilize the BitShift operator instead of MatMul during inference, which effectively reduces hardware resource requirement. 
Experimental results show that our fully quantized vision transformers achieve comparable performance with the full precision models.
Altogether, we provide a higher baseline for future works, hoping that FQ-ViT’s strong performance will encourage research into even lower bit-width quantization of vision transformers, which will boost their real-world adoptions.

\clearpage
\newpage
\small{\bibliographystyle{named}
\bibliography{ijcai22}}

\newpage
\appendix
\input{supp_text}

\end{document}

%% file: supp_text.tex
\section{Quantized Inference}
\subsection{Quantized Inference of LayerNorm}
LayerNorm has been widely used in neural networks to accelerate the convergence during the training process, which can be formulated as:
\begin{align}
    \label{ln}
     \operatorname{LayerNorm}(\textrm{X}) &= \frac{\textrm{X}-\mu_\textrm{X}}{\sqrt{\sigma_\textrm{X}^2+\epsilon}}\cdot\gamma + \beta\\
    &=\frac{\gamma}{\sqrt{\sigma_\textrm{X}^2+\epsilon}}\textrm{X} + \frac{\beta\sqrt{\sigma_\textrm{X}^2+\epsilon} - \gamma\mu_\textrm{X}}{\sqrt{\sigma_\textrm{X}^2+\epsilon}},
\end{align}

where $\gamma, \beta$ are the learned parameters and $\mu_\textrm{X},\sigma_\textrm{X}$ are the statistics which need to be calculated based on the input of LayerNorm.

In this paper, proposed Power-of-Two Factor is used for LayerNorm input $\textrm{X}\in \mathds{R}^{\textrm{B}\times \textrm{L}\times \textrm{C}}$ and its quantized value $\textrm{X}_\textrm{Q}$ can be written as:

\begin{equation}
    \textrm{X}_\textrm{Q}=\textrm{Q}(\textrm{X}|b)=\operatorname{clip}(\lfloor\frac{\textrm{X}}{\textrm{2}^{\alpha}s}\rceil+zp, \textrm{0},\textrm{2}^{b}-1),
\end{equation}
where $s,zp\in \mathds{R}^\textrm{1}$ are quantization parameters and $\alpha \in \mathds{N}^\textrm{C}$ is Power-of-Two Factor.

Following the LayerNorm's definition, we should calculate the statistics of input $\textrm{X}$. As described in this paper, the whole process can be divided into two phases. In the first phase, we shift the quantized activation $\textrm{X}_\textrm{Q}$ with PTF $\alpha$:
\begin{equation}
    \widehat{\textrm{X}}_\textrm{Q}=(\textrm{X}_\textrm{Q}-zp)<<\alpha.
\end{equation}
Then, in the second phase, we need to calculate the statistics based on the shifted activation $\widehat{X}_\textrm{Q}$. Firstly, we gauge the mean of $\textrm{X}$ and $\textrm{X}^\textrm{2}$ as follows:
\begin{align}
    \mu_\textrm{X}&\approx\frac{1}{\textrm{C}}\sum_{i=1}^{\textrm{C}}(\widehat{\textrm{X}}_{\textrm{Q}i}\cdot s)=\frac{s}{\textrm{C}}\sum_{i=1}^{\textrm{C}}(\widehat{\textrm{X}}_{\textrm{Q}i})\rightarrow\frac{s}{\textrm{C}}\textrm{M}_1,\\
    \mu_{\textrm{X}^2}&\approx\frac{1}{\textrm{C}}\sum_{i=1}^{\textrm{C}}(\widehat{\textrm{X}}_{\textrm{Q}i}\cdot s)^2=\frac{s^2}{\textrm{C}}\sum_{i=1}^{\textrm{C}}(\widehat{\textrm{X}}_{\textrm{Q}i})^2\rightarrow\frac{s^2}{\textrm{C}}\textrm{M}_2,
\end{align}
where $\textrm{C}$ is the number of channels in $\textrm{X}$.
Secondly, we utilize $\mu_\textrm{X}$ and $ \mu_{\textrm{X}^2}$ to calculate $\sigma_\textrm{X}^2$:
\begin{equation}
    \sigma_\textrm{X}^2=\mu_{\textrm{X}^2} -\mu_{\textrm{X}}^2\approx \frac{s^2}{\textrm{C}^2}(\textrm{C}\textrm{M}_2-\textrm{M}_1^2),
\end{equation}
and approximate $\sqrt{\sigma_\textrm{X}^2+\epsilon}$ as:
\begin{equation}
    \sqrt{\sigma_\textrm{X}^2+\epsilon}\approx  \frac{s}{\textrm{C}}\sqrt{\textrm{C}\textrm{M}_2-\textrm{M}_1^2}.
\end{equation}

Thus, we obtain the statistics of input $\textrm{X}$ based on integer-only calculations.

After the calculations of statistics, we need to do the integrated inference of LayerNorm.
Equation~(\ref{ln}) can be written as follows by quantizing the input $\textrm{X}$ and output $\textrm{Y}_\textrm{Q}$:
\begin{equation}
\textrm{Y}_\textrm{Q}=\lfloor\frac{ s_{in}\gamma}{d_{out}\sqrt{\sigma_\textrm{X}^2+\epsilon}}\widehat{\textrm{X}}_\textrm{Q} +\frac{\beta\sqrt{\sigma_\textrm{X}^2+\epsilon}-\gamma\mu_\textrm{X}}{s_{out}\sqrt{\sigma_\textrm{X}^2+\epsilon}}\rceil+{zp}_{out},
\end{equation}
where $s_{in}\in \mathds{R}^1$ is the scale of input $\textrm{X}$, and $s_{out},{zp}_{out}\in \mathds{R}^1$ are the scale of output $\textrm{Y}_\textrm{Q}$.

In order to simplify the equation, we fuse each term as follows:
\begin{align}
    \textrm{A}&=\frac{s_{in}\gamma}{s_{out}\sqrt{\sigma_\textrm{X}^2+\epsilon}},\\
    \textrm{B}&=\frac{\beta\sqrt{\sigma_\textrm{X}^2+\epsilon}-\gamma\mu_\textrm{X}}{s_{out}\sqrt{\sigma_\textrm{X}^2+\epsilon}}.
\end{align}
To obtain the integer-only inference, we approximate $\textrm{A}$ as:
\begin{equation}
    \textrm{N}_1= b - 1 - \lfloor \log_2|\textrm{A}|\rfloor, \textrm{N}_2=\lfloor |\textrm{A}|  2^{\textrm{N}_1}\rfloor,
\end{equation}
\begin{equation}
    \textrm{A}=\operatorname{sign}(\textrm{A}) \cdot \frac{\textrm{N}_2}{2^{\textrm{N}_1}},
\end{equation}
where $b$ is target bit-width.
Finally, the quantized inference for LayerNorm can be formulated as:
\begin{align}
    \textrm{Y}_\textrm{Q}&=\lfloor \textrm{A}\widehat{\textrm{X}}_\textrm{Q}+\textrm{B}\rceil+{zp}_{out}\\
    &=\lfloor \frac{\operatorname{sign}(\textrm{A}) \cdot\textrm{N}_2\widehat{\textrm{X}}_\textrm{Q} + \lfloor \textrm{B}\cdot 2^{\textrm{N}_1}\rceil }{2^{\textrm{N}_1}}\rceil +{zp}_{out}.
\end{align}

\subsection{Quantized Inference of Softmax}
\begin{figure}[t]
    \centering
    \includegraphics[width=8.2cm]{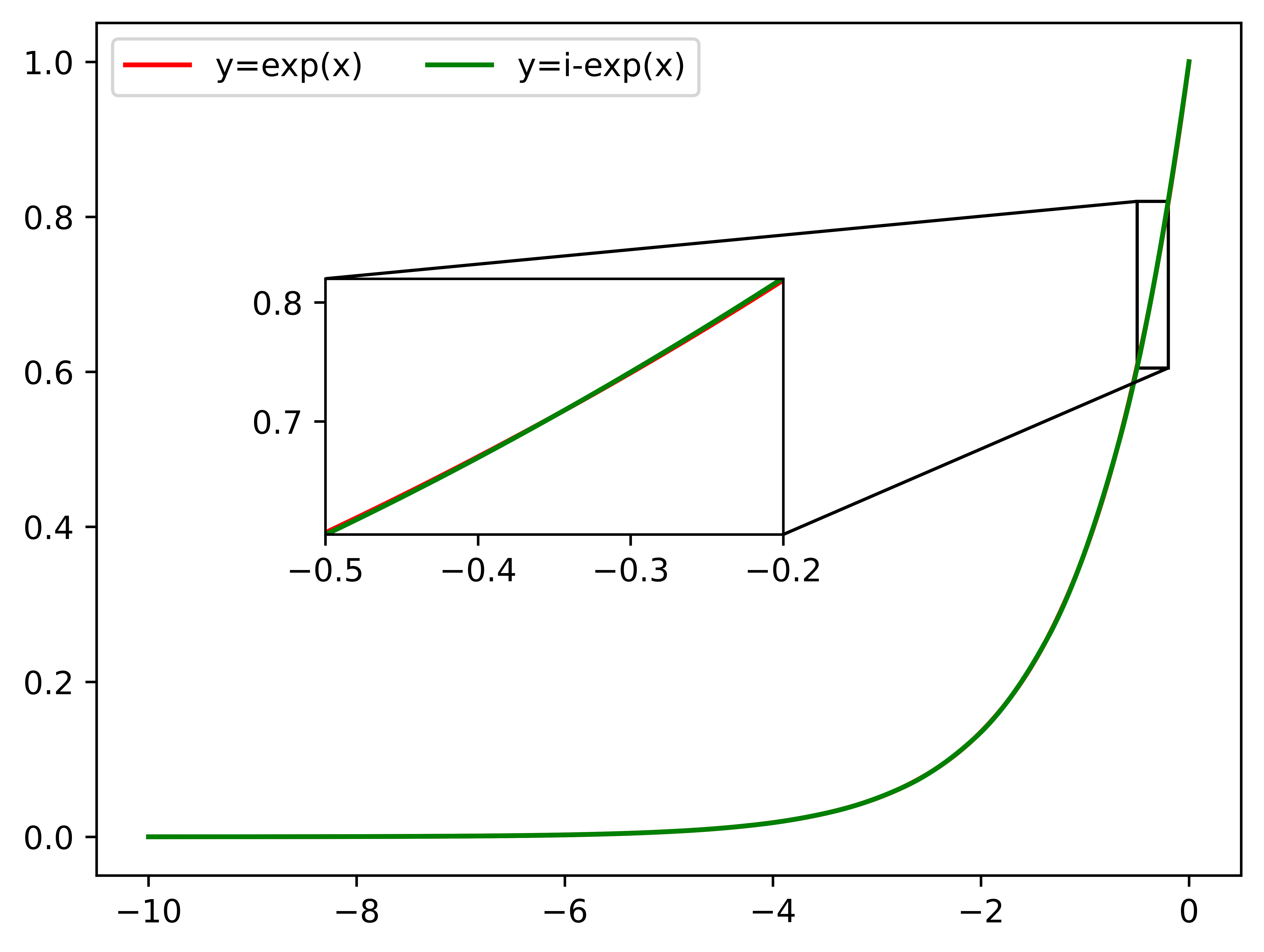}
    \caption{Comparison between exponential~(exp) and integer-only exponential~(i-exp)}
    \label{iexp_vs}
\end{figure}
Softmax is generally used in the Vision Transformers and it can be formulated as:
\begin{equation}
    \operatorname{Softmax}(\textrm{X})_i = \frac{\exp(\textrm{X}_i)}{\sum_{j=1}^{J}\exp(\textrm{X}_j)}.
\end{equation}

In practice, we subtract the maximum value of the input to avoid overflowing:
\begin{equation}
    \tilde{\textrm{X}}_i = \textrm{X}_i-\operatorname{max}(\textrm{X}),
\end{equation}
and thus the Softmax can be written as:
\begin{equation}
    \operatorname{Softmax}(\textrm{X})_i = \frac{\exp(\tilde{\textrm{X}}_i)}{\sum_{j=1}^{J}\exp(\tilde{\textrm{X}}_j)}.
\end{equation}

Now, all inputs, $\tilde{\textrm{X}}$, become non-positive. We decompose it as $(-\ln 2)z+p$, where $z$ is a non-negative integer and the $p$ is a real number in $(-\ln 2, 0]$. Then, the exponential of $\tilde{\textrm{X}}$ can be written as:
\begin{equation}
    \exp(\tilde{\textrm{X}}) = 2^{-z}\exp(p) = \exp(p)>>z,
\end{equation}
where $>>$ is the right shifting.

I-BERT use a second-order polynomial to approximate the exponential function in the interval of $p\in (-\ln2, 0]$:
\begin{equation}
    L(p) = 0.3585(p+1.353)^2+0.344\approx\exp(p),
\end{equation}
\begin{equation}
    \exp(\tilde x)\approx \operatorname{i-exp}(\tilde x):= L(p)>>z,
\end{equation}
where $z = \lfloor -\tilde{\textrm{X}} / \ln2\rfloor$ and $p = \tilde{\textrm{X}} + z\ln2$.

Figure~\ref{iexp_vs} plots the result of $\operatorname{i-exp}$, which is nearly identical to the exponential function. The largest gap between these two functions is only $1.9\times 10^{-3}$, which is relatively negligible comparing to 8-bit quantization error of $1/255 = 3.9 \times 10^{-3}$.

\begin{algorithm}[t]
\caption{Integer-only Exponential}
\label{iexp-alg}
\begin{algorithmic}[1]
\STATE {\bfseries Input:} $q,s$: quantized input and scale
\STATE {\bfseries Output:} $q_{out},s_{out}$: quantized output and scale
\vskip 0.075in
\FUNCTION{\textsc{I-Exp}{$(q, s)$}}
{
\STATE $q_{\ln2}\leftarrow\lfloor-\ln2/s\rfloor$
\STATE $q\leftarrow max(q, n\cdot q_{\ln2})$
\STATE $z\leftarrow\lfloor q/q_{\ln2}\rfloor$
\STATE $q_{p}\leftarrow q-z\cdot q_{\ln2}$
\STATE $q_b, q_c \leftarrow \lfloor b/s\rfloor, \lfloor c/as^2\rfloor$
\STATE $s_L\leftarrow \lfloor as^2\rfloor$
\STATE $q_L = (q+q_b)^2+q_c$
\STATE $q_{out}, s_{out}\leftarrow q_L<<n-z, s_L/2^n$
\STATE \Return $q_{out}, s_{out}$
}
\ENDFUNCTION
\end{algorithmic}
\end{algorithm}

\begin{algorithm}[t]
\caption{Log-Int-Softmax}  
\label{lis-alg}
\begin{algorithmic}[0] 
\STATE {\bfseries Input:} $q,s$: quantized input and input scale
\STATE {\bfseries Output:} $q_{out},s_{out}$: quantized output and output scale
\vskip 0.075in
\FUNCTION{\textsc{I-Log2}$(q)$}
{
\STATE $ \textrm{M} \leftarrow $ $\operatorname{Find\_First\_One}(q)$
\STATE $ \chi \leftarrow (\textrm{M}-1)^{th}$ Bit of $q$
\STATE $q_{out}\leftarrow \textrm{M}+\chi $
\STATE \Return $q_{out}$
}
\ENDFUNCTION
\FUNCTION{\textsc{Log-Int-Softmax}$(q, s)$}{
\STATE $\tilde q \leftarrow q-\operatorname{max}(q)$
\STATE $q_{exp}, s_{exp}\leftarrow \operatorname{I-Exp}(\tilde q, s)$
\STATE $q_{rev} \leftarrow \lfloor \operatorname{sum}(q_{exp})/q_{exp}\rceil$ 
\STATE $q_{out}, s_{out} \leftarrow \textrm{N} - \operatorname{I-Log2}(q_{rev}), 1/2^N$
\STATE \Return $q_{out}, s_{out}$
}
\ENDFUNCTION
\end{algorithmic}  
\end{algorithm}  

Based on $\operatorname{i-exp}$~(Algorithm~\ref{iexp-alg}), we propose Log-Int-Softmax~(LIS) as Algorithm~\ref{lis-alg} shows. Firstly, the maximum value of inputs is subtracted to ensure all the inputs are non-positive, and the results are sent to the $\operatorname{i-exp}$. After that, we replace the original Softmax by its reciprocal to ensure the results of integer division to be larger than 1. Last but not least, we perform a Log2 quantization on the output of reverse Softmax as:
\begin{align}
      \operatorname{LIS}(s\cdot \textrm{X}_\textrm{Q})&= \textrm{N} - \operatorname{clip}(\log_2\lfloor\frac{\sum \operatorname{i-exp}(\textrm{X}_\textrm{Q})}{\operatorname{i-exp}(\textrm{X}_\textrm{Q})}\rceil, 0, 2^b-1)\\
      &= \textrm{N}-\textrm{Attn}_\textrm{Q},
     \label{log2_equation}
\end{align}
with $\textrm{N}=2^b-1$. $\textrm{X}_\textrm{Q}, s$ are the quantized input and scale. 

The Log2 function can be calculated with integer arithmetic as showed in Algorithm \ref{lis-alg}. To obtain the Log2 of the integer, we introduce the $\operatorname{Find\_First\_One}$ function which returns the index of the most significant one bit of the input. The whole process can be done in integer domain.
For example, if the input of $\operatorname{I-Log2}$ is $0000\ 1101\ 1010\ 1100_{2}$, the $\textrm{M}, \chi$ will be calculated as $11,1$ and the rounding result will be 12.

The following step of Softmax is the matrix multiplication between attention map $\textrm{Attn}$ and values $\textrm{V}$. Considering $\textrm{V}_\textrm{Q}$ is the quantized value of $V$ and its scale and zero point are $s_\textrm{V},{zp}_\textrm{V}\in \mathds{R}^1$, the matrix multiplication can be written as:
\begin{align}
    \textrm{Attn}\cdot \textrm{V} &= \frac{2^{\textrm{N}-\textrm{Attn}_\textrm{Q}}}{2^\textrm{N}}\cdot (\textrm{V}_\textrm{Q}-{zp}_\textrm{V})s_\textrm{V}\\
    & = \frac{s_\textrm{V}}{2^\textrm{N}} \cdot (\textrm{V}_\textrm{Q}-{zp}_\textrm{V}) << (\textrm{N}-\textrm{Attn}_\textrm{Q}).
\end{align}
As we can see, the multiplication is converted to BitShift operator with an output scale $\frac{s_\textrm{V}}{2^\textrm{N}}$, where $\textrm{N}$ equals $2^4-1=15$ in 4-bit quantization.

Based on the above processes, we realize fully fixed-point inference for Softmax and accelerate the calculation of attention map and values by using BitShift.

\section{Hyperparameter \textrm{K} of Power-of-Two Factor}
\begin{figure}[t]
\centering
\includegraphics[width=8.3cm]{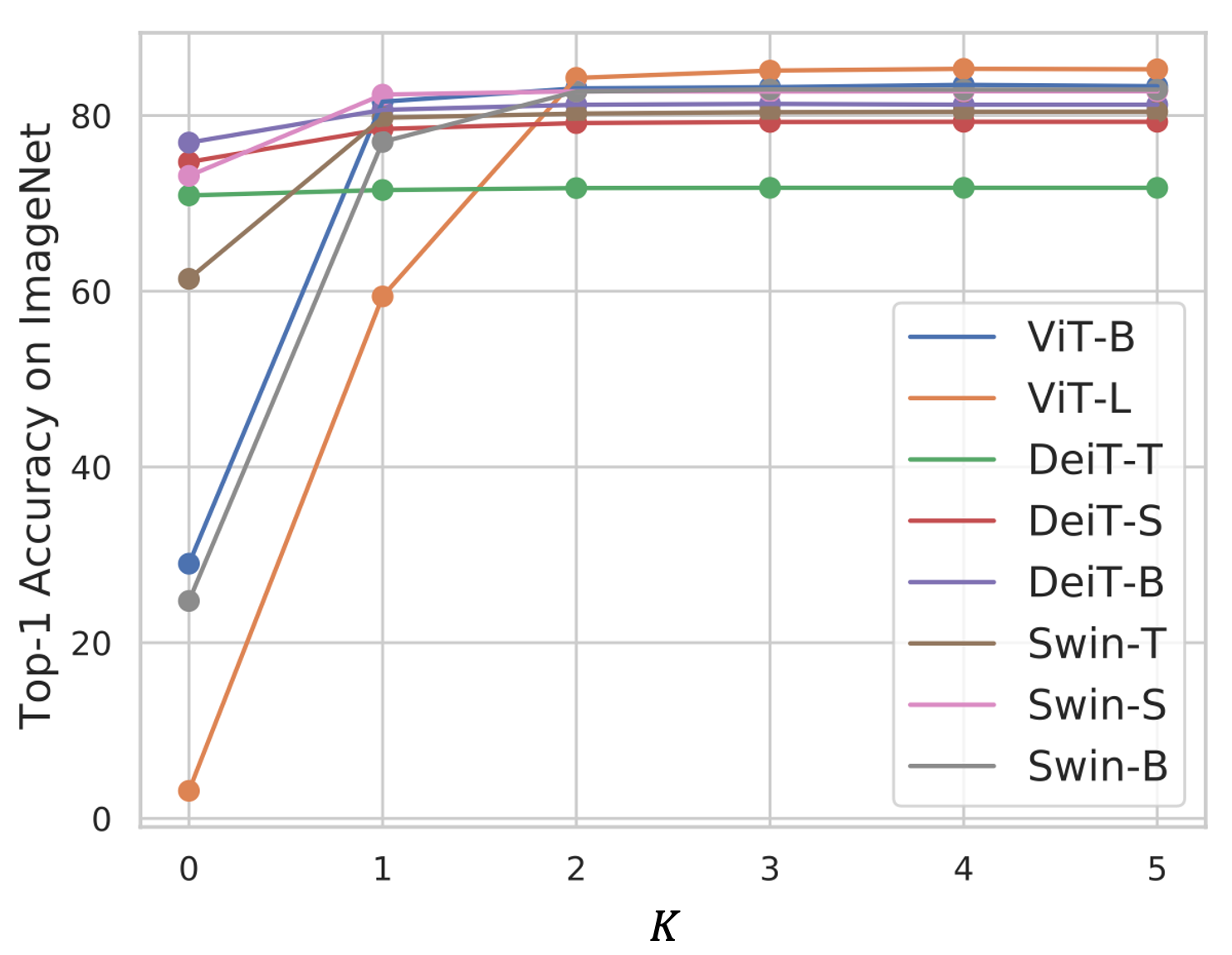}
\caption{Top-1 accuracy on ImageNet with different hyperparameter K.}
\label{K}
\end{figure}

We conduct experiments on the setting of hyperparameter \textrm{K} in PTF. As shown in Figure~\ref{K}, we evaluate the top-1 accuracy of vision transformers with different \textrm{K}. It is observed that the accuracy is nearly saturated when \textrm{K} is taken as 3. As a result, to meet the different inter-channel variants between all models as much as possible, we choose 3 as the default value.

\section{Inter-channel Variation for Vision Transformers and ResNets}
\begin{figure*}[t]
    \centering
    \includegraphics[width=15cm]{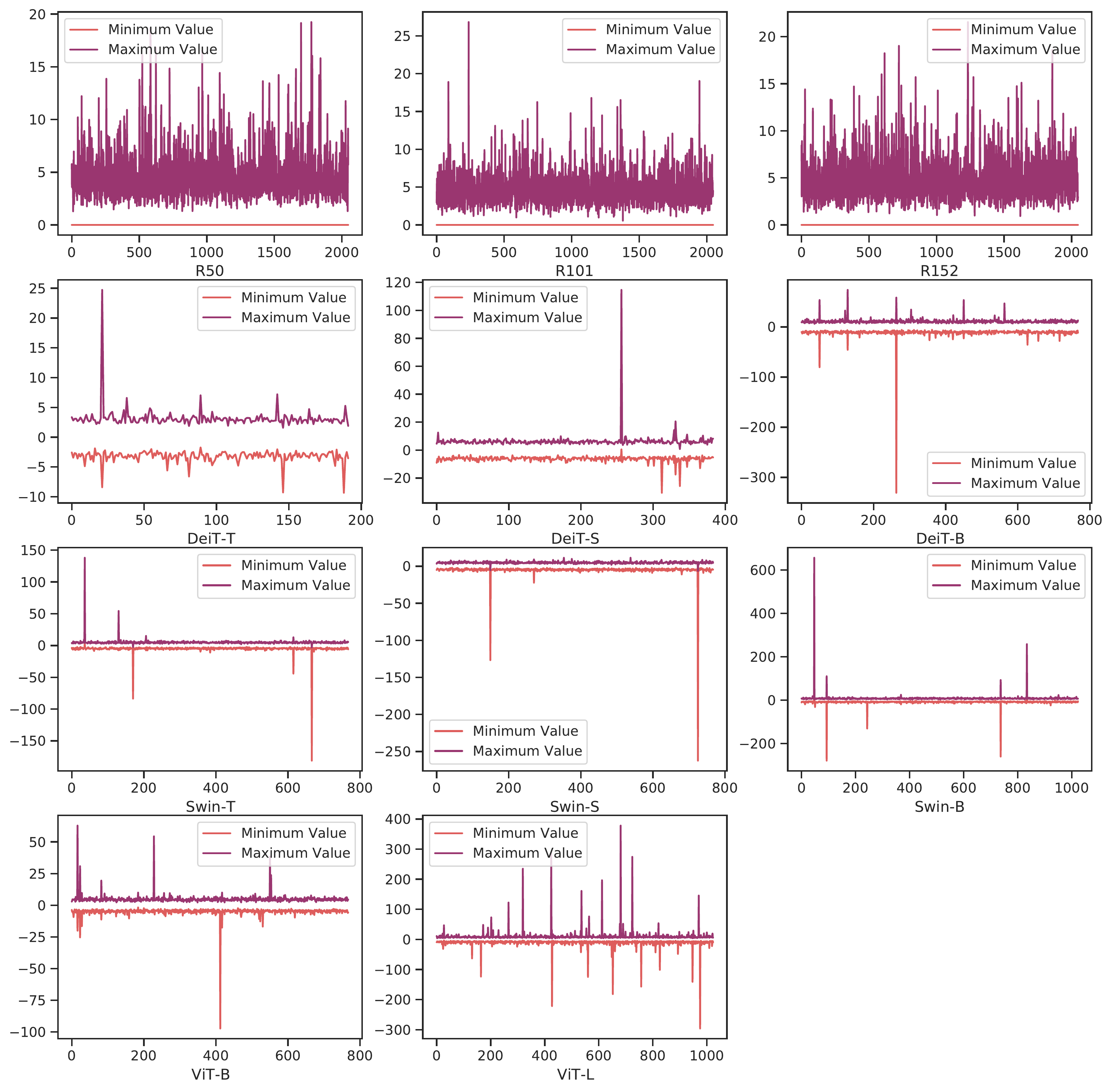}
    \caption{Channel-wise minimum and maximum values of Vision Transformers and ResNets.}
    \label{channel}
\end{figure*}
In Figure~\ref{channel}, we present the channel-wise minimum and maximum values of Vision Transformers and ResNets. For comparison, we choose the input of the last LayerNorm layer for Vision Transformers and the output of the fourth stage for ResNets to show. It is observed that a serious inter-channel variation are found in Vision Transformers.